\documentclass{article}

\usepackage{microtype}
\usepackage{microtype}
\usepackage{graphicx}
\usepackage{subcaption}
\usepackage{booktabs}
\usepackage{nicefrac}
\usepackage{natbib}
\usepackage{doi}
\usepackage{arxiv}
\usepackage{xspace}
\usepackage{caption}

\usepackage{amsmath}
\usepackage{amssymb}
\usepackage{mathtools}
\usepackage{amsthm}
\usepackage{hyperref}
\hypersetup{
  colorlinks=true,
  linkcolor=black,
  citecolor=blue,
  urlcolor=blue,
  hypertexnames=false,
  pdfborderstyle={}
}
\usepackage[all]{hypcap}
\usepackage[capitalize,noabbrev]{cleveref}

\newcommand{\methodname}{GATES\xspace}

\theoremstyle{plain}

\theoremstyle{definition}

\theoremstyle{remark}

\usepackage{enumitem}
\setlist{nosep}

\usepackage[textsize=tiny]{todonotes}

\fancypagestyle{firstpage}{%
  \fancyhf{}%
  \lfoot{\footnotesize *Corresponding author: \texttt{astein0@umd.edu}}%
}


\begin{document}

% \title{\methodname: Self-Distillation Under Asymmetric Context Information}
% \title{\methodname: Gated Asymmetric Trajectory Self-Distillation}
\title{\methodname: Self-Distillation under Privileged Context with Consensus Gating}

% \author{
%   \large Alex Stein \\
%   % Department of Computer Science\\
%   University of Maryland, College Park\\
%   \texttt{astein0@umd.edu} \\
%   \And
%   \large Furong Huang \\
%   % Department of Computer Science\\
%   University of Maryland, College Park\\
%   \And
%   \large Tom Goldstein \\
%   % Department of Computer Science\\
%   University of Maryland, College Park\\
% }

\author{
  \large Alex Stein\footnotemark \\
  % Department of Computer Science\\
  University of Maryland, College Park\\
  \And
  \large Furong Huang \\
  % Department of Computer Science\\
  University of Maryland, College Park\\
  \And
  \large Tom Goldstein \\
  % Department of Computer Science\\
  University of Maryland, College Park\\
}

\date{}

\maketitle

\renewcommand{\shorttitle}{GATES: Self-Distillation under Privileged Context with Consensus Gating}

\thispagestyle{firstpage}

\begin{abstract}
We study self-distillation in settings where supervision is unreliable: there are no ground truth labels, verifiable rewards, or external graders to evaluate answers.
We focus on document-grounded question answering with asymmetric context, where a single model serves as both tutor (with access to a relevant source document during training) and student (answering from the question alone at test time).
Rather than assuming tutor correctness, we derive supervision online from tutor consensus by sampling multiple document-grounded reasoning traces and using agreement to gate learning.
Conditioned on this reliability signal, we distill knowledge through full tutor reasoning trajectories (not just final answers), providing a dense and stable learning signal.
Empirically, this consensus-gated trajectory distillation substantially improves transfer to the document-free student.
Held-out in-domain accuracy under asymmetric evaluation improves from 46.0\% to 62.0\%, and average (maj@8) accuracy on public document-free math benchmarks improves from 20.2\% to 35.4\%.
\end{abstract}

\begin{figure}[b]
    \centering
    \includegraphics[width=\linewidth]{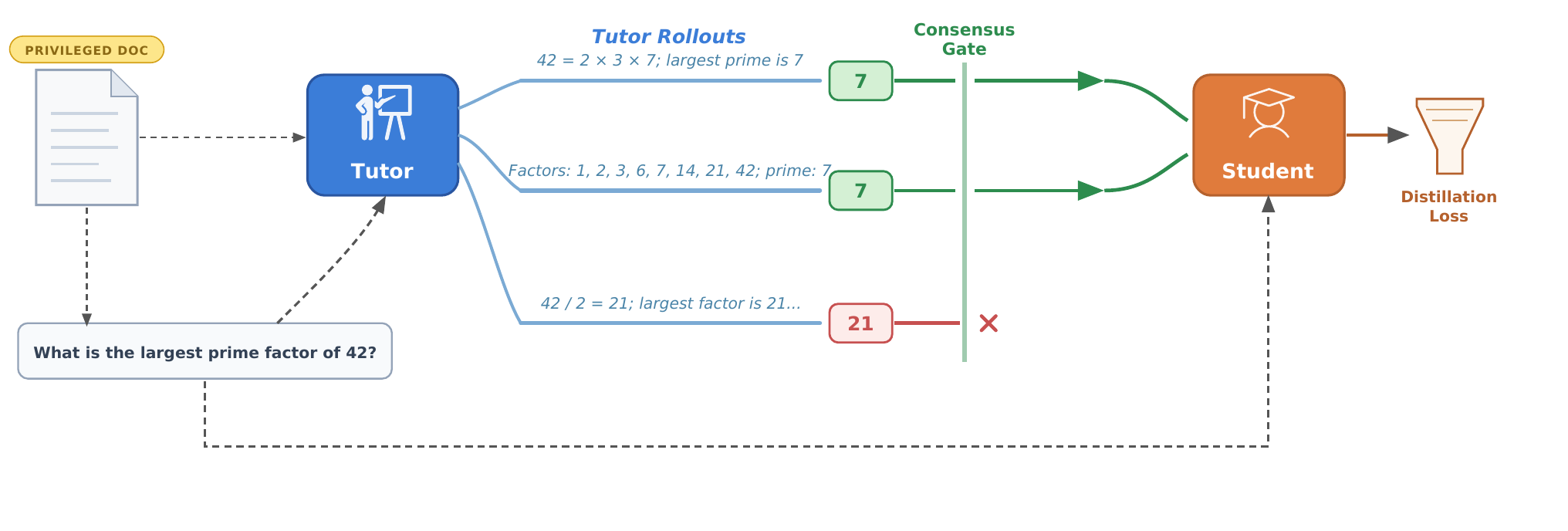}
    \caption{Overview of \methodname. A tutor model, given a privileged document and question, generates multiple reasoning rollouts. A consensus gate filters rollouts based on answer agreement, discarding trajectories with minority answers. The surviving rollouts are used to train a student model---which receives only the question---via distillation loss, transferring the tutor's privileged reasoning without requiring ground-truth labels. The tutor and student share the same underlying model, differing only in whether the privileged document is included in the input context.}
    \label{fig:overview}
\end{figure}

\section{Introduction}

Language model fine-tuning often relies on ground-truth labels, verifiable rewards, or an external grader to evaluate answers \citep[e.g.,][]{cobbe2021trainingverifierssolvemath, lightman2023letsverifystepstep, uesato2022solvingmathwordproblems}.
In the absence of such supervision, distillation is appealing because it provides dense, token-level supervision and can transfer knowledge through full reasoning trajectories, not just final answers \citep{zelikman2022starbootstrappingreasoningreasoning, yuan2025selfrewardinglanguagemodels, singh2024humandatascalingselftraining}.
However, standard distillation assumes an asymmetry between teacher and student, where a larger or more capable teacher provides reliable supervision that the student learns to imitate.
In self-distillation, the teacher \emph{is} the model itself, so naive self-distillation can reinforce systematic errors or encourage degenerate shortcuts that are easy to imitate.

We study a setting in which a model can self-improve without any verified labels or external grader.
Consider a question generated from a reference document; for example, a word problem derived from a passage containing relevant facts or examples.
A model prompted \emph{with} the document can reason over the relevant evidence and is more likely to answer correctly; the same model prompted \emph{without} the document must rely on its own internalized knowledge and is less reliable.
We call these two roles the \emph{tutor} (with document) and the \emph{student} (without document), but they are the same model with the same weights, differing only in whether the document is included in the input.
This asymmetry in input context creates the imbalance that standard distillation obtains from a larger or more capable teacher, making self-distillation meaningful.

The tutor is not an oracle. 
Because its only advantage lies in its ability to perform reasoning over the reference document, it remains susceptible to errors.
To avoid distilling from incorrect supervision, we sample $k$ independent tutor responses during training and only learn from questions where at least $\tau$ responses agree on the same final answer.
When this consensus is strong, we distill the full tutor reasoning trajectories (not only final answers) into the document-free student, providing dense token-level supervision.
When consensus is weak, the question is skipped entirely and contributes zero loss.
Unlike self-consistency \citep{wang2023selfconsistencyimproveschainthought}, which selects answers at inference time from a single context, our consensus signal gates \emph{training} updates across an asymmetric context gap---it decides \emph{when} to distill, not \emph{what} to answer.

Empirically, our method (which we call \methodname{}) substantially outperforms all baselines.
Held-out in-domain student accuracy improves from 46.0\% to 62.0\%, and average maj@8 accuracy on public document-free math benchmarks improves from 20.2\% to 35.4\% (\S\ref{sec:results}; Figure~\ref{fig:main_results}).
Naive alternatives fail in this regime: answer-only supervised fine-tuning catastrophically degrades student accuracy to 10–12\%, and reward-only optimization provides no meaningful improvement over the pretrained base model (Figure~\ref{fig:main_results}a).

Our contributions are:
\begin{itemize}
    \item We formalize asymmetric-context self-distillation, a setting in which a single model serves as both tutor and student under different input contexts, enabling self-improvement without ground-truth labels, verifiable rewards, or an external grader.
    \item We introduce \methodname{} (\textbf{G}ated \textbf{A}symmetric \textbf{T}rajectory \textbf{S}elf-distillation), which uses agreement among multiple document-grounded tutor responses to gate which reasoning trajectories are distilled into the student.
    \item We show empirically that consensus gating is the critical mechanism: \methodname{} outperforms unfiltered trajectory distillation, answer-only fine-tuning, and outcome-based RL, and ablations confirm that removing the gate alone accounts for the performance gap (\S\ref{sec:ablations}).
\end{itemize}
% \input{chapters/problem_setting}
% ----------------
\section{Related Work}

\paragraph{Contextual and Asymmetric Distillation}
Classical knowledge distillation transfers behavior from a teacher to a student, typically in an off-policy fashion where the student is trained on teacher-generated trajectories \citep{hinton2015distillingknowledgeneuralnetwork,gou2021knowledge,sun2019patientknowledgedistillationbert}.
Prior work has studied on-policy variants that better match the student distribution and can improve stability \citep{onpolicydistillationlanguagemodels,thinkingmachines_onpolicy_distillation_blog}.
Our setting is also related to learning with privileged information and weak-to-strong capability transfer, where additional information may be available at training time but not at test time \citep{openai_weak_to_strong_generalization}.
The notion of asymmetric information at training and test time was formalized as \emph{learning using privileged information} (LUPI) by \citet{VAPNIK2009544}; our tutor-student setup instantiates this framework with document access as the privileged modality.
We differ from standard distillation in two key ways: (i) the tutor is not an external or reliably correct oracle, and (ii) the tutor has access to privileged context unavailable to the student.
Rather than assuming correctness, we infer supervision reliability online via consensus among multiple document-grounded tutor rollouts.
Document-grounded question answering is a well-studied setting \citep{kwiatkowski2019naturalquestionsbenchmarkquestion, yang2018hotpotqadatasetdiverseexplainable}; we use it not as an end goal but as a testbed where asymmetric access to evidence arises naturally.

\paragraph{Self-Training, Self-Distillation, and Consensus}
Self-training and self-distillation approaches improve a model using its own generations \citep{he2020revisiting}, but a recurring challenge is error amplification, where naive self-training reinforces its own mistakes.
A common mitigation is to filter or refine pseudo-labels using self-consistency and related signals \citep{madaan2023selfrefineiterativerefinementselffeedback,bai2022constitutionalaiharmlessnessai}.
Majority voting and consensus-based filtering are well-established reliability mechanisms in this literature; they have been used for answer selection at inference time \citep{wang2023selfconsistencyimproveschainthought}, for filtering training data in self-improvement pipelines \citep{zelikman2022starbootstrappingreasoningreasoning}, and for offline reinforcement learning on self-generated data \citep{gulcehre2023reinforcedselftrainingrest}.
Our use of tutor consensus plays the same role: it gates learning updates based on agreement among tutor rollouts.
The key difference is that our consensus signal is derived from tutor agreement under asymmetric context rather than from verified correctness labels and is used to gate dense trajectory-level distillation rather than to filter correct-answer chains.

Preference-based self-improvement methods offer a related but 
distinct approach, learning from internally generated comparisons \citep{rafailov2024directpreferenceoptimizationlanguage,azar2023generaltheoreticalparadigmunderstand,wu2024selfplaypreferenceoptimizationlanguage}; these approaches assume the model can generate meaningfully contrastive pairs, whereas our method derives supervision from agreement rather than preference.

\paragraph{Concurrent Work on Self-Distillation}
Several recent works, appearing concurrently with ours, apply self-distillation to reasoning but under different supervision assumptions. Some assume access to reliable correctness signals: \citet{zhao2026selfdistilledreasoneronpolicyselfdistillation} stabilize on-policy RL by distilling improved trajectories back 
into the policy, while \citet{qu2026popelearningreasonhard} use privileged information to guide exploration with correctness feedback.
Both methods rely on verified correctness signals to determine which trajectories to learn from; our setting removes this assumption entirely, using only self-agreement under asymmetric context as the supervision signal. 
Others target complementary settings: \citet{shenfeld2026selfdistillationenablescontinuallearning} treat self-distillation as a regularization mechanism for continual learning, and \citet{hübotter2026reinforcementlearningselfdistillation} 
reinterpret RL itself as self-distillation, relying on preference signals to define trajectory quality.
Our work is distinguished by the combination of asymmetric input context (the tutor and student are the same model but receive different information) and unreliable supervision, where tutor consensus is the sole mechanism for deciding when distillation is trustworthy.

\paragraph{Self-Play and Question Generation}
Self-play methods generate training data by adaptively constructing tasks that challenge the learner, often using reinforcement learning updates.
This paradigm has a long history in game-playing agents, where self-play alone (i.e., without human data) proved sufficient to achieve superhuman performance \citep{silver2017masteringgamegowithout, silver2018generalreinforcementlearningalgorithm}.
Recent work has extended self-play and data-free training to language model reasoning and curriculum construction \citep{liu2025spiceselfplaycorpusenvironments,kuba2025languageselfplaydatafreetraining,zhao2025absolutezeroreinforcedselfplay,huang2026rzeroselfevolvingreasoningllm}.
In this work, most experiments use a fixed challenger (offline question generation) in order to isolate the learning dynamics of consensus-gated distillation.
We view adaptive challenger optimization as complementary rather than competing, and present preliminary evidence in \S\ref{sec:discussion}.

% \input{chapters/method}
% =====================================
% §3 METHOD — COMBINED (old §2 + old §4)
% =====================================
\section{\methodname: Gated Asymmetric Trajectory Self-Distillation}\label{sec:method}

We now formalize the asymmetric-context setting and describe the gated self-distillation training procedure.
A single model $\pi_\theta$ operates as both tutor (conditioned on document $d$ and question $q$) and student (conditioned on $q$ alone). 
We derive supervision online from tutor consensus and distill full reasoning trajectories into the document-free student.

% --------------------------------------------------
\subsection{Setting and Notation}\label{sec:setting}
% --------------------------------------------------

Let $\mathcal{C}$ denote a corpus of documents, and let $d \in \mathcal{C}$ be the document associated with a question $q$.
We use a fixed set of questions pregenerated from the corpus, without verified answers during training. 
These questions can, in principle, be generated adaptively online, as in self-play systems~\citep{liu2025spiceselfplaycorpusenvironments,kuba2025languageselfplaydatafreetraining,zhao2025absolutezeroreinforcedselfplay,huang2026rzeroselfevolvingreasoningllm}, but our method does not require adaptive question generation; in this work, we use a \emph{fixed challenger} that pregenerates a dataset of document–question pairs from $\mathcal{C}$ before training begins (we pregenerate one question per document).  Details on the entire question generation procedure in \S\ref{sec:setup}.

The \emph{student} is the model evaluated at test time, and the \emph{tutor} is the same model instance (sharing all parameters) queried with additional context.
The only difference between the two roles is their prompt: the tutor is conditioned on $(d, q)$, while the student is conditioned on $q$ alone.
Because both roles share the same weights, any improvement from training benefits the model as a whole; the student and tutor are updated simultaneously during training, with no separate tutor model or delayed copy.
We assume that no verified answers are available during training.
Instead, learning is driven entirely by tutor sampling and a consensus gate, which is described in the next section.

% --------------------------------------------------
\subsection{Consensus-Gated Training}\label{sec:consensus_training}
% --------------------------------------------------

The central mechanism of \methodname{} is \emph{consensus gating}: for each training question, we sample $k$ independent tutor rollouts and extract a final answer from each.
We declare \emph{strong consensus} when at least $\tau$ of the $k$ rollouts agree on the same extracted answer $a^*$; otherwise, the question is skipped entirely and contributes zero loss.
When consensus is strong, we treat the modal answer $a^*$ as a pseudo-label (notably, $a^*$ is the most frequent tutor answer, not a ground-truth label) and distill eligible tutor trajectories into the document-free student.

Formally, each training step proceeds over a batch of $n$ document–question pairs $(d_i, q_i)$.
For each question, we generate $k$ tutor rollouts and $k$ student rollouts.
We then:
\begin{enumerate}
    \item \textbf{Gate by consensus.} Extract a final answer (the last \verb|\boxed{...}| expression) from each tutor rollout and compute a question-level consensus gate $g_i \in \{0,1\}$, where $g_i = 1$ if at least $\tau$ of $k$ tutor answers agree and $g_i = 0$ otherwise. 
    \item \textbf{Filter by eligibility.} For each rollout $j$, compute a rollout-level eligibility indicator $e_{i,j} \in \{0,1\}$, where $e_{i,j} = 1$ if the rollout's extracted answer matches the consensus answer $a^*_i$ and the rollout passes document-leakage guardrails (keyword filtering for explicit document references; see Appendix~\ref{sec:guardrails_appendix}).
    \item \textbf{Update.} Compute the training losses defined in \S\ref{sec:objectives} using only questions with $g_i = 1$ and, for off-policy losses, only rollouts with $e_{i,j} = 1$.
\end{enumerate}
When $g_i = 0$, the question contributes zero loss across all objectives.
This operationalizes our core idea: learning updates occur only when supervision is inferred to be reliable, and learning is performed at the token level rather than only on final answers.

\begin{figure}[!t]
  \centering
  \includegraphics[width=\textwidth]{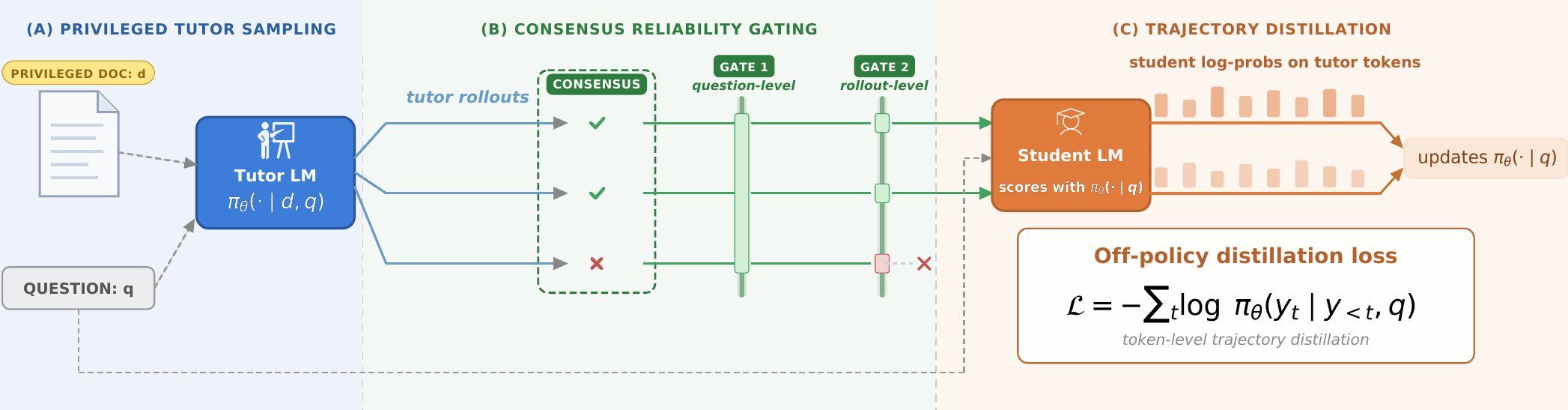}
    \caption{Off-policy distillation in \methodname{}. A single model $\pi_\theta$ operates in two roles under asymmetric context: as a \emph{tutor} conditioned on both the source document $d$ and question $q$, and as a \emph{student} conditioned on $q$ alone. \textbf{(A)}~The tutor generates $k$ independent reasoning rollouts per question. \textbf{(B)}~A question-level consensus gate labels the question as \emph{reliable} if sufficiently many rollouts agree on the same answer; a second, rollout-level gate retains only trajectories that match the consensus. \textbf{(C)}~Eligible tutor trajectories provide dense token-level supervision to the document-free student via trajectory distillation (Eq.~1). Unreliable questions are skipped entirely, preventing self-reinforcement collapse.}
  \label{fig:off-policy-schematic-final}
\end{figure}

\begin{figure}[!t]
  \centering
  \includegraphics[width=\textwidth]{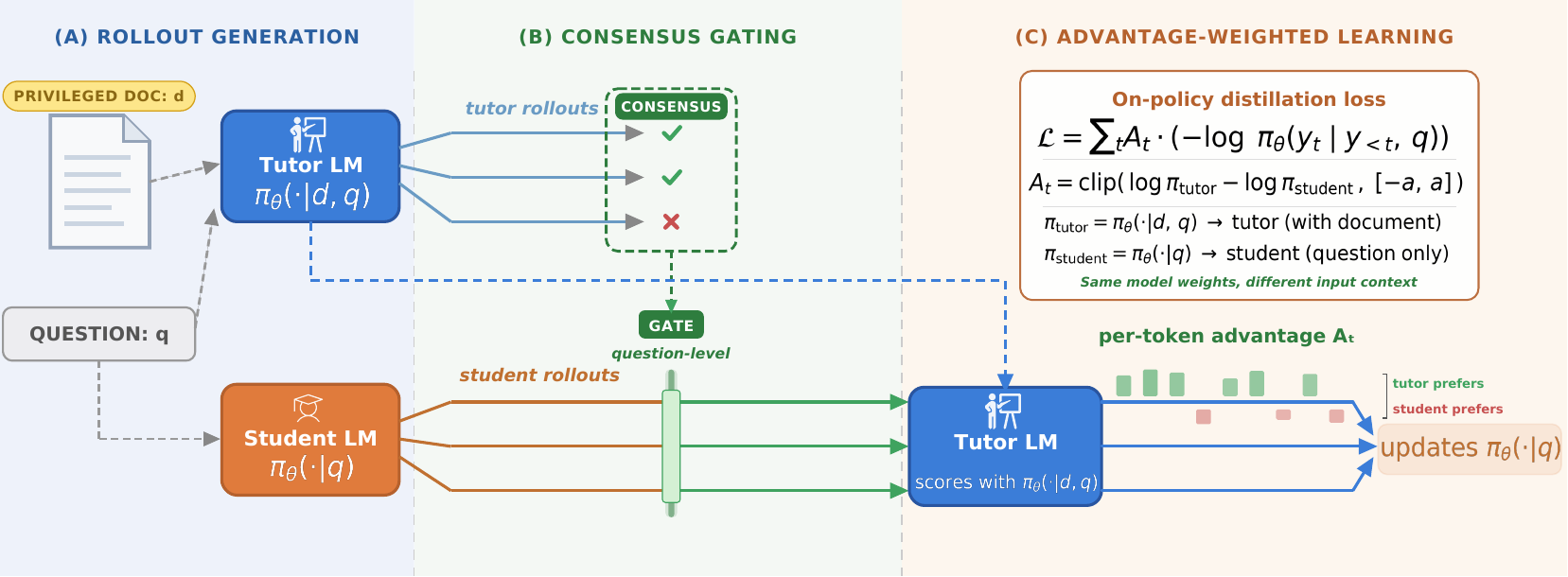}
    \caption{On-policy distillation in \methodname{}. As in the 
      off-policy variant (Figure~\ref{fig:off-policy-schematic-final}), 
      a single model $\pi_\theta$ serves as both \emph{tutor} and 
      \emph{student} under asymmetric context. \textbf{(A)}~Both 
      roles generate $k$ rollouts in parallel: tutor rollouts 
      establish consensus, while student rollouts provide the 
      on-policy training signal. \textbf{(B)}~A question-level 
      consensus gate determines reliability; unlike the off-policy 
      setting there is no trajectory-level filter---when consensus 
      is strong, all student rollouts pass through. 
      \textbf{(C)}~The tutor scores each token of the student's 
      own rollouts by computing log-probabilities under document 
      context. The per-token advantage 
      $A_t = \mathrm{clip}(\log \pi_{\mathrm{tutor}} - \log 
      \pi_{\mathrm{student}},\, [-a,a])$ upweights tokens where 
      the tutor assigns higher probability, encouraging 
      document-grounded reasoning while remaining on-policy 
      (Eq.~3). Unreliable questions contribute zero loss.}
  \label{fig:on-policy-schematic-final}
\end{figure}

\paragraph{On-policy vs.\ off-policy distillation}
\methodname{} supports two complementary modes of distillation.
In \emph{off-policy} distillation, the student directly imitates eligible tutor-generated trajectories (Figure~\ref{fig:off-policy-schematic-final}).
In \emph{on-policy} distillation, the student generates its own trajectories, and the tutor scores each token by computing log-probabilities under document context; the resulting per-token advantage upweights tokens where the document-aware tutor assigns higher probability than the student (Figure~\ref{fig:on-policy-schematic-final}).
Both modes are gated by the same consensus mechanism.
In practice, off-policy trajectory-level distillation provides the primary performance gains, while on-policy updates contribute modest additional improvement (\S\ref{sec:ablations}).

% --------------------------------------------------
\subsection{Training Objectives}\label{sec:objectives}
% --------------------------------------------------

We optimize the student policy $\pi_\theta(\cdot \mid q)$ using two dense distillation losses, both gated by the consensus signal described in \S\ref{sec:consensus_training}.
We write $\pi_T(\cdot \mid d, q)$ for the tutor distribution and $\pi_\theta(\cdot \mid q)$ for the student distribution.
To avoid overloading $T$ (which denotes the tutor), we use $L$ for trajectory length; $L^{(T)}_{i,j}$ and $L^{(S)}_{i,j}$ denote the number of completion tokens in the $j$-th tutor and student rollout for question $i$, respectively.

\paragraph{Off-policy distillation (tutor rollouts)}
Let $y^{(T)}_{i,j} = \left(y^{(T)}_{i,j,1},\ldots,y^{(T)}_{i,j,L^{(T)}_{i,j}}\right)$ denote the $j$-th tutor completion (the full reasoning trajectory including the final answer) for question $i$.
The off-policy distillation loss (negative log-likelihood on tutor tokens) is
\begin{equation}
\begin{aligned}
\mathcal{L}_{\text{off}}(\theta)
&= -\frac{1}{\sum_{i,j} g_i\,e_{i,j}\,L^{(T)}_{i,j}}
\sum_{i,j} g_i\,e_{i,j}
\sum_{t=1}^{L^{(T)}_{i,j}} \ell^{(T)}_{i,j,t}(\theta),\\
\qquad \ell^{(T)}_{i,j,t}(\theta) &\coloneqq \log \pi_\theta\!\left(y^{(T)}_{i,j,t}\,\mid\, y^{(T)}_{i,j,<t},q_i\right).
\end{aligned}
\end{equation}
Only tutor trajectories whose final extracted answer matches the consensus and that pass the guardrails are included.

\paragraph{On-policy distillation (student rollouts)}
On-policy distillation uses student-generated trajectories weighted by an advantage-like signal.
Let $y^{(S)}$ be a student completion sampled from $\pi_\theta(\cdot \mid q)$.
We compute:
\begin{equation}
\begin{aligned}
A_t &= \text{clip}\Bigl(\log \pi_T\bigl(y^{(S)}_t\,\mid\, y^{(S)}_{<t},d,q\bigr)\\
&\quad- \log \pi_\theta\bigl(y^{(S)}_t\,\mid\, y^{(S)}_{<t},q\bigr),\,[-a,a]\Bigr),
\end{aligned}
\end{equation}
where $a$ is a clipping hyperparameter, and gradients do not flow through $A_t$.
The on-policy loss is:
\begin{equation}
\begin{aligned}
\mathcal{L}_{\text{on}}(\theta)
&= -\frac{1}{\sum_{i,j} g_i\,L^{(S)}_{i,j}}
\sum_{i,j} g_i
\sum_{t=1}^{L^{(S)}_{i,j}} A_{i,j,t}\,\ell^{(S)}_{i,j,t}(\theta),\\
\qquad \ell^{(S)}_{i,j,t}(\theta) &\coloneqq \log \pi_\theta\!\left(y^{(S)}_{i,j,t}\,\mid\, y^{(S)}_{i,j,<t},q_i\right).
\end{aligned}
\end{equation}
This upweights tokens that the document-aware tutor assigns higher likelihood than the student, encouraging document-grounded reasoning while staying on-policy.

\paragraph{Total objective}
The training objective combines both distillation losses:
\begin{equation}
\mathcal{L}(\theta) = \lambda_{\text{off}}\,\mathcal{L}_{\text{off}} + \lambda_{\text{on}}\,\mathcal{L}_{\text{on}}.
\end{equation}
This two-term objective can be extended with auxiliary losses.
We define a consensus-correctness reward ($\mathcal{L}_{\text{cons}}$) that treats tutor agreement as a sparse binary REINFORCE signal; we ablate this term in \S\ref{sec:ablations}.
We also apply a KL regularization term toward a frozen reference policy to prevent drift from the base model.
Both auxiliary terms are formally defined in Appendix~\ref{sec:auxiliary_losses}.
% =====================================
% §5 EXPERIMENTS — REVISED
% =====================================
\section{Empirical Evaluation}\label{sec:experiments}

\subsection{Experimental Setup}\label{sec:setup}

\paragraph{Model and dataset}
We use Qwen3-4B-Base as the underlying model for all experiments.
We construct a fixed-challenger dataset by prompting Qwen2.5-32B-Instruct to generate questions from documents in the Nemotron-CC-Math corpus~\citep{nemotron}, following a procedure similar to SPICE~\citep{liu2025spiceselfplaycorpusenvironments}. 
% \fhc{Shall we talk about this? Let's do something novel here.}
For each candidate question, we generate $k{=}8$ tutor rollouts with document context and require at least $5/8$ tutor answers to agree; questions that fail this strict consensus filter are dropped.
Validity additionally requires producing a parsable final answer (the last \verb|\boxed{...}| expression) and satisfying document-leakage guardrails (Appendix~\ref{sec:guardrails_appendix}).
The resulting dataset contains 551 training questions and 50 held-out evaluation questions (Table~\ref{tab:dataset_sizes}).
All experiments use this fixed challenger; adaptive curricula are discussed in \S\ref{sec:discussion}.

\begin{table}[!h]
\centering
\small
\begin{tabular}{lr}
\toprule
Split & \# Questions \\
\midrule
Train & 551 \\
Held-out eval & 50 \\
\bottomrule
\end{tabular}
\caption{Dataset sizes for the fixed-challenger splits.}
\label{tab:dataset_sizes}
\end{table}

\paragraph{Training details}
Unless otherwise noted, we use a batch size of $n{=}32$ questions per step, $k{=}8$ tutor rollouts per question, and $k{=}8$ student rollouts per question.
We train for a single epoch over the training split.
The consensus gate requires $\geq 4/8$ agreeing tutor extracted answers (\S\ref{sec:consensus_training}).
Both tutor and student are prompted to produce a step-by-step solution terminating in a \verb|\boxed{...}| answer; we extract the last boxed expression for consensus and evaluation.
All losses are computed on completion tokens only (prompt tokens are masked), and distillation losses are normalized by the total number of included completion tokens after masking.
Extracted answers are compared using the \texttt{math-verify} library~\citep{math_verify}.
For a complete set of hyperparameters, see Appendix~\ref{sec:hyperparameters}.

\paragraph{Evaluation}
We report accuracy under symbolic final-answer equivalence.
For in-domain evaluation, we measure student accuracy on the held-out split (50 questions), and, where applicable, tutor accuracy on the same questions with document access.
For this evaluation only, correctness is measured against an oracle answer produced by Qwen2.5-32B-Instruct using a consensus over 8 rollouts (temperature ${=}0.3$, requiring $\geq 5/8$ agreement); this oracle is used only for evaluation.
For out-of-domain evaluation, we report student accuracy on four public document-free math benchmarks(MATH, AMC, Minerva, OlympiadBench) using majority voting over 8 samples (maj@8).
This metric is a natural fit for \methodname{}:
consensus gating trains the model to learn only from self-consistent trajectories,
so an agreement-based decoding strategy directly reflects the consistency the method optimizes for.

\paragraph{Baselines}
We compare against four training signals that isolate the roles of dense imitation and sparse correctness.
\textbf{Answer-Only SFT} fine-tunes Qwen3-4B-Base on question--answer pairs using the challenger-provided extracted answer (the \verb|\boxed{...}| expression only, not the full reasoning trajectory) as the target.
\textbf{Answer-Only SFT (w/ doc)} fine-tunes on document--question--answer triplets using the tutor prompt, again targeting only the extracted answer.
\textbf{Outcome RL} trains the model with a REINFORCE policy gradient using a binary correctness reward derived from tutor consensus, plus KL regularization toward the pretrained reference; no trajectory-level distillation is performed.
This isolates the effect of dense trajectory-level distillation from the underlying self-generated supervision signal.
\textbf{Tutor-Trajectory SFT} fine-tunes on the full tutor reasoning trajectories (including chain-of-thought and final answer) without consensus gating, training on all tutor rollouts regardless of agreement. 
This isolates the contribution of the gating mechanism by providing the same dense trajectory signal without reliability filtering.
% The tutor--student gap for this baseline is especially revealing(Figure~\ref{fig:main_results}(b)): Tutor-Trajectory SFT achieves the highest tutor accuracy (74\%) but only 54\% student accuracy.
This baseline exhibits a wide tutor--student gap: high tutor accuracy but substantially lower student accuracy.
Without gating, the student internalizes reasoning patterns that implicitly depend on document access; the tutor can recover from flawed intermediate steps by re-reading the source, but the student cannot.
Consensus gating filters out these document-dependent trajectories, ensuring the student learns only from reasoning that is self-sufficient.
See Appendix~\ref{app:additional_eval} for further analysis of the tutor performance.

% ------------------------------------------------------------------
\subsection{Main Results}\label{sec:results}
% ------------------------------------------------------------------

\begin{figure}[t]
\centering

\small
\begin{tabular}{lccccc}
  \toprule
  \textbf{Method} & \textbf{MATH} & \textbf{AMC} & \textbf{Minerva} & \textbf{OlympiadBench} & \textbf{Avg.} \\
  \midrule
  \methodname{} (ours) & \textbf{65.6} & \textbf{26.5} & \textbf{21.7} & \textbf{27.7} & \textbf{35.4} \\
  \midrule
  Tutor-Trajectory SFT & 60.8 & 24.1 & 19.5 & 24.7 & 32.3 \\
  Outcome RL   & 44.0 & 10.8 & 18.4 & 12.0 & 21.3 \\
  Base Model   & 42.6 & 9.6 & 15.8 & 12.6 & 20.2 \\
  Answer-Only SFT  & 35.6 & 8.4 & 13.2 & 7.7 & 16.2 \\
  Answer-Only SFT (w/ doc) & 35.8 & 10.8 & 14.3 & 7.3 & 17.1 \\
  \bottomrule
\end{tabular}

\vspace{2pt}
\textbf{(a)} Document-free math benchmarks (maj@8)

\vspace{8pt}
\includegraphics[width=0.85\textwidth]{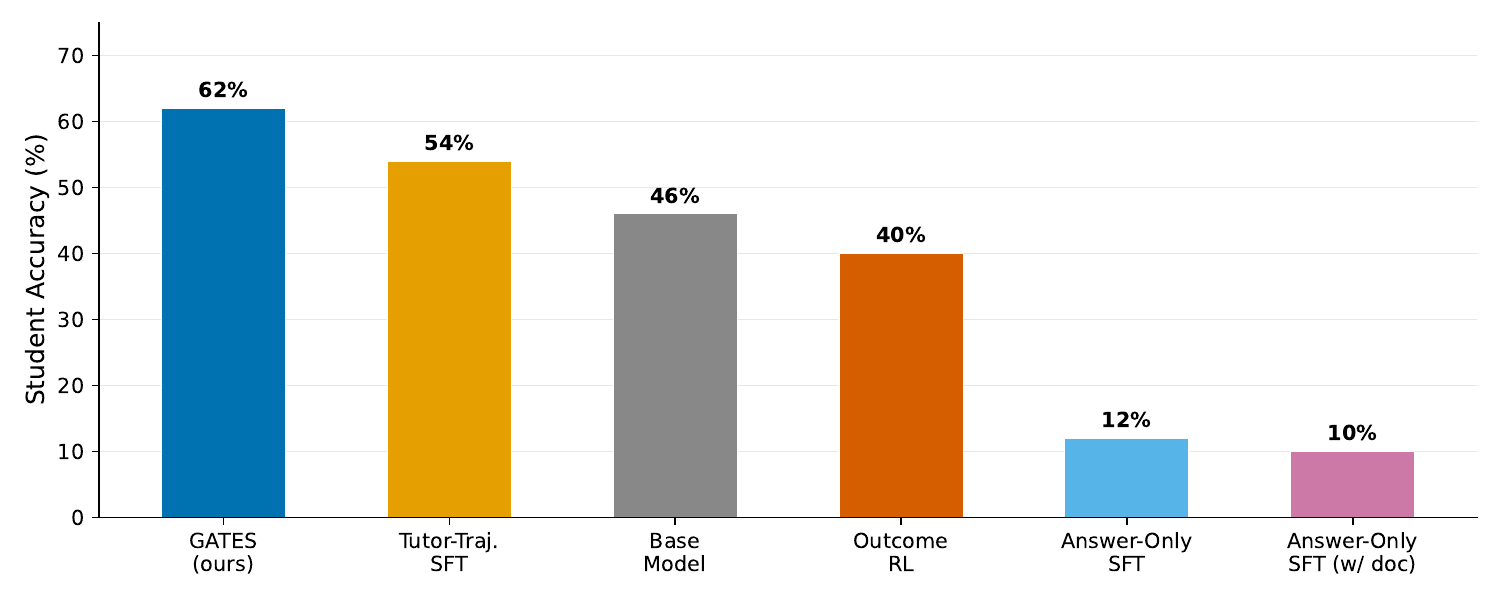}

\vspace{2pt}
\textbf{(b)} In-domain asymmetric evaluation

\caption{Main results comparing \methodname{} against baselines.
(a)~Accuracy (\%) on four document-free math benchmarks (maj@8 decoding).
(b)~Student accuracy on the held-out asymmetric evaluation (50 questions, greedy decoding).
\methodname{} yields the best student accuracy (62\%), improving 16 percentage points over the pretrained base model.
See Appendix~\ref{app:additional_eval} for tutor accuracy, greedy decoding, and coverage results.}
\label{fig:main_results}
\end{figure}

We evaluate whether \methodname{} improves a document-free student without verified labels.
Figure~\ref{fig:main_results}(b) reports held-out in-domain accuracy under asymmetric evaluation, which directly reflects the supervision setting we study.
Figure~\ref{fig:main_results}(a) reports accuracy on public document-free math benchmarks as a complementary stress test.

On document-free benchmarks (Figure~\ref{fig:main_results}(a)), \methodname{} outperforms all baselines, improving average maj@8 accuracy from 20.2\% (pretrained) to 35.4\%.
Tutor-Trajectory SFT, which trains on all tutor rollouts without consensus gating, is the next strongest baseline (32.3\%), but still falls 3.1 percentage points short of \methodname{}, confirming that reliability filtering provides a meaningful benefit even when full reasoning trajectories are available.
Outcome RL and the pretrained base model achieve comparable benchmark averages (21.3\% and 20.2\%, respectively), indicating that sparse reward feedback alone provides no meaningful improvement.

The held-out in-domain evaluation (Figure~\ref{fig:main_results}(b)) provides the strongest evidence of effective transfer.
\methodname{} improves student accuracy from 46.0\% to 62.0\%,
demonstrating effective transfer from document-grounded supervision.
Both Answer-Only SFT baselines catastrophically degrade student accuracy (to 12.0\% and 10.0\%).
This confirms that naive fine-tuning on extracted answers alone (without full reasoning trajectories) destroys reasoning capability and that consensus-gated trajectory distillation is essential for stable learning.
The tutor outperforms the student by 34.3 percentage points on the filtered training set (70.1\% vs.\ 35.8\%), confirming that document access provides a substantial advantage that \methodname{} successfully transfers.

% ------------------------------------------------------------------
\subsection{Ablations}\label{sec:ablations}
% ------------------------------------------------------------------

\begin{figure}[t]
\centering

\small
\begin{tabular}{lcccccccc}
  \toprule
  \textbf{Configuration} & $\lambda_{\text{off}}$ & $\lambda_{\text{on}}$ & $\lambda_{\text{cons}}$ & \textbf{MATH} & \textbf{AMC} & \textbf{Minerva} & \textbf{OBench} & \textbf{Avg.} \\
  \midrule
  \methodname{} (ours)  & 1.0 & 0.1 & 0.0 & \textbf{65.6} & 26.5          & 21.7          & \textbf{27.7} & \textbf{35.4}          \\
  $+$ Oracle Loss        & 1.0 & 0.1 & 1.0 & 64.2          & \textbf{27.7} & \textbf{23.5} & 27.4          & \textbf{35.7} \\
  $-$ Gate              & 1.0 & 0.1 & 0.0 & 59.2          & 21.7          & 18.4          & 25.0          & 31.1          \\
  Oracle Only          & 0.0 & 0.0 & 1.0 & 60.8          & 25.3          & 23.2          & 23.4          & 33.2          \\
  All Losses           & 1.0 & 1.0 & 1.0 & 61.0          & 20.5          & 21.7          & 25.8          & 32.2          \\
  On-Policy Dominant   & 0.1 & 1.0 & 1.0 & 60.2          & 19.3          & 21.3          & 21.6          & 30.6          \\
  On-Policy + Oracle   & 0.0 & 1.0 & 1.0 & 58.6          & 15.7          & 20.2          & 20.0          & 28.6          \\
  \midrule
  Base Model           & --  & --  & --  & 42.6          & 9.6           & 15.8          & 12.6          & 20.2          \\
  \bottomrule
\end{tabular}

\vspace{2pt}
\textbf{(a)} Benchmark ablations (maj@8)

\vspace{8pt}
\includegraphics[width=0.85\textwidth]{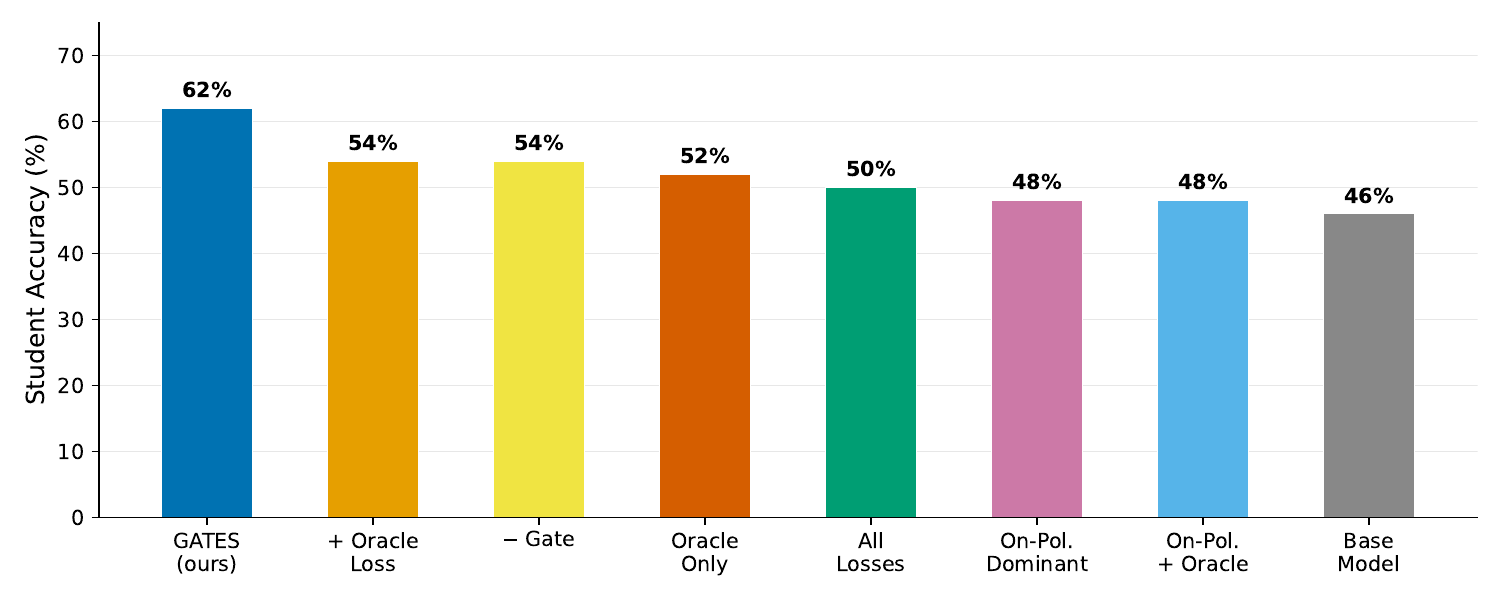}

\vspace{2pt}
\textbf{(b)} In-domain ablations 

\caption{Ablation results varying loss weights with $\lambda_{\text{KL}} = 0.02$ fixed throughout.
(a)~Accuracy (\%) on document-free math benchmarks (maj@8 decoding).
\emph{\methodname{} (ours)} is the canonical configuration.
$-$~Gate uses the same loss weights but removes consensus gating, resulting in a $4.3$ pp benchmark drop despite identical loss configuration.
Adding oracle loss provides no meaningful improvement (35.7 vs.\ 35.4), confirming that consensus gating alone is sufficient without verified correctness labels.
(b)~Student accuracy on the held-out asymmetric evaluation (greedy decoding).
Off-policy distillation remains the dominant contributor: configurations without it show the largest drops in student accuracy.
Tutor accuracy is reported in Appendix~\ref{app:additional_eval}.}
\label{fig:ablations}
\end{figure}

We ablate which components of the training objective are necessary for reliable transfer.
Figure~\ref{fig:ablations}(a) and Figure~\ref{fig:ablations}(b) report benchmark and in-domain results across seven configurations that vary the loss weights while holding all other hyperparameters fixed.

Off-policy distillation and consensus gating are the dominant contributors to transfer.
The canonical \methodname{} configuration ($\lambda_{\text{off}}{=}1.0$, $\lambda_{\text{on}}{=}0.1$, $\lambda_{\text{cons}}{=}0.0$) achieves the strongest student accuracy (62.0\%) and a benchmark average of 35.4\%.
Removing the consensus gate while keeping the same loss weights ($-$~Gate) drops student accuracy to 54.0\% and benchmark average to 31.1\%, isolating the contribution of reliability filtering.
Notably, $-$~Gate achieves nearly identical results to Tutor-Trajectory SFT (31.1\% vs.\ 32.3\% benchmark average, both 54\% student accuracy), confirming that consensus gating (rather than other aspects of the training pipeline) is the active ingredient distinguishing \methodname{} from unfiltered trajectory distillation.
Reversing the off-policy and on-policy weights (On-Policy Dominant) drops student accuracy to 48.0\% and benchmark average to 30.6\%, confirming that trajectory-level imitation of tutor rollouts is the primary mechanism of transfer.
Removing off-policy distillation entirely (On-Policy + Oracle) yields similar degradation.

Adding the consensus-correctness reward ($\mathcal{L}_{\text{cons}}$) does not meaningfully change performance.
The ``+ Oracle Loss'' configuration achieves a comparable benchmark average (35.7\% vs.\ 35.4\%), but reduces student accuracy (54.0\% vs.\ 62.0\%), suggesting that the sparse correctness signal does not complement consensus-gated distillation.
The Oracle~Only configuration (sparse reward with no trajectory-level distillation) achieves moderate results (student 52.0\%, benchmark avg.\ 33.2\%), confirming that dense trajectory imitation adds meaningful value beyond sparse correctness feedback.

% ------------------------------------------------------------------
\subsection{Summary and Discussion of Results}\label{sec:results_summary}
% ------------------------------------------------------------------

\methodname{} produces the strongest student performance in our experiments, both on held-out in-domain evaluation and on public math benchmarks.
Off-policy distillation from tutor rollouts is the primary driver of transfer; on-policy updates provide modest additional improvement when anchored by consensus-based reliability gating.
Together, these results establish the minimality of the \methodname{} mechanism: the only component required beyond standard off-policy self-distillation is consensus gating.
The oracle correctness reward is unnecessary (\S\ref{sec:ablations}), the explicit consistency loss adds no benefit ($\lambda_{\text{cons}}{=}0$), and no external teacher or reward model is involved.
Consensus gating alone provides sufficient reliability modeling to make self-distillation under unreliable supervision effective.
Naive alternatives (supervised fine-tuning and outcome-based reinforcement learning) fail to produce meaningful improvements, underscoring the importance of dense trajectory-level supervision gated by explicit reliability modeling.

After training, the student more consistently produces structured solutions that terminate in a parsable final answer, even though it never sees the document.
The document-leakage guardrails (\S\ref{sec:consensus_training}) are essential for preventing the student from learning to reference evidence it will not have at test time.

Additional evaluation under greedy decoding and coverage 
(pass@8) is reported in Appendix~\ref{app:additional_eval}.
% ----------------
\section{Discussion}\label{sec:discussion}
% Purpose: explain why this worked at all.

\subsection{Why the Method Works}
Our method succeeds by explicitly separating two requirements that are often conflated in self-training: \emph{reliability} (a signal that correlates with correctness) and \emph{learnability} (a dense objective that can be optimized stably).
Asymmetric context creates a reliability gap: because the tutor has access to the source document during training, it can condition on evidence that the student will not see, inducing a systematic tutor–student gap that makes distillation meaningful.
We use tutor agreement as a reliability test: if multiple document-grounded rollouts converge to the same answer, we treat that instance as trustworthy enough to learn from.
On those trusted instances, we distill the full trajectory, which provides the student with dense token-level supervision and avoids the variance of sparse reward updates.
Empirically, \methodname{}'s consensus-gated trajectory distillation is sufficient: the sparse correctness reward ($\mathcal{L}_{\text{cons}}$) provides no additional benefit in our setting, while dense imitation without consensus gating underperforms (\S\ref{sec:ablations}).
Interestingly, adding the oracle correctness reward is neutral on benchmarks but reduces in-domain student accuracy (54\% vs. 62\%).
One possible explanation is that the oracle reward upweights correct-but-low-consensus trajectories that the gate would otherwise filter, partially reintroducing document-dependent reasoning that the student cannot replicate without the source.

\subsection{Limitations}

\methodname{} relies on agreement among multiple tutor rollouts as a proxy for correctness.
If the tutor is insufficiently capable, biased, or produces low-diversity reasoning traces, consensus may reflect shared errors rather than reliable supervision.
Relatedly, to avoid reinforcing incorrect supervision, we discard all questions without sufficient tutor agreement, which improves reliability but reduces the effective number of training updates and may limit sample efficiency.

Our approach also assumes that final answers can be reliably extracted and normalized.
If answer extraction is noisy (or if tasks lack a well-defined final answer), both consensus estimation and downstream learning can degrade.
Estimating tutor consensus further requires multiple tutor rollouts per question, increasing training-time computation relative to single-pass supervision.

Finally, we evaluate exclusively in document-grounded question answering, where privileged document access provides a natural and strong asymmetry.
If the privileged context provides little additional evidence beyond what is already implied by the question, or if the tutor is not systematically more reliable than the student on the training distribution, then there may be little useful signal to transfer.

\subsection{Future Directions}

\begin{figure}[!t]
\centering
\includegraphics[width=\textwidth]{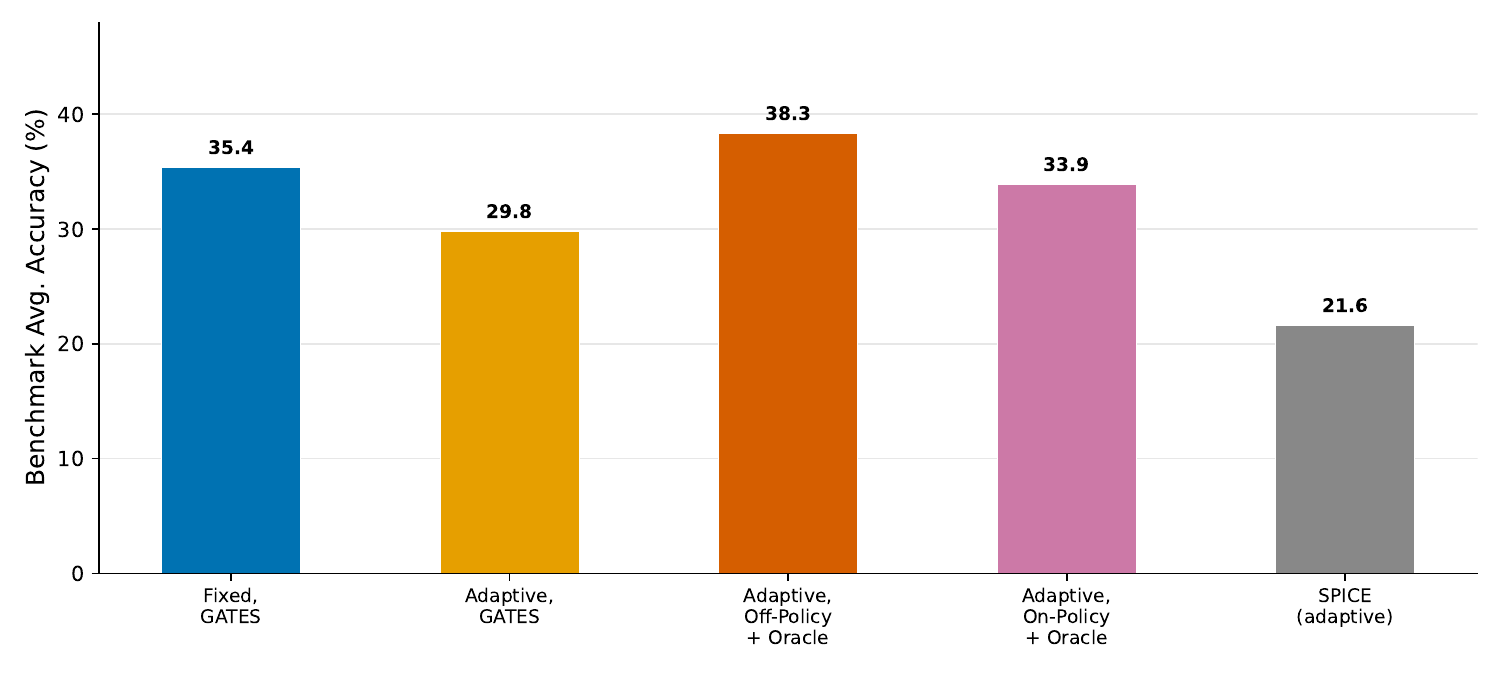}
\caption{Benchmark average accuracy (\%) under fixed vs.\ adaptive challenger training (maj@8 decoding). 
\emph{Fixed, \methodname{}} and \emph{Adaptive, \methodname{}} use the canonical configuration ($\lambda_{\text{off}}\!=\!1.0$, $\lambda_{\text{on}}\!=\!0.1$, $\lambda_{\text{cons}}\!=\!0.0$) with a static or adaptive challenger, respectively.
The remaining adaptive variants add the oracle loss ($\lambda_{\text{cons}}\!=\!1.0$) under different on/off-policy weightings. 
SPICE uses an adaptive challenger by design and is included for reference under a matched update budget.}
\label{fig:adaptive}
\end{figure}

\paragraph{Adaptive challengers}
A natural extension of \methodname{} is to generate questions on the fly using an adaptive challenger, potentially improving coverage and curriculum quality.
Unlike the fixed challenger, which uses Qwen2.5-32B-Instruct to pre-generate questions offline, adaptive training generates questions from the model itself, making it a strictly harder setting with no external question source.
Preliminary experiments (Figure~\ref{fig:adaptive}) suggest that adaptive training can improve out-of-distribution benchmark performance, with the best adaptive variant reaching 38.3\% average accuracy compared to 35.4\% under the fixed challenger.
Notably, the configuration that performs best under adaptive training differs from the canonical \methodname{} setup: adding the oracle loss ($\mathcal{L}_{\text{cons}}$) appears to help when questions are generated adaptively, possibly because harder or less familiar questions increase the prevalence of confident but incorrect tutor agreement (the primary failure mode of consensus gating), and the oracle loss provides a direct corrective signal for exactly these cases.
Without this grounding, the adaptive \methodname{} configuration reaches 29.8\% (below the fixed challenger (35.4\%) but still well above the pretrained baseline (20.2\%)), suggesting that the optimal loss composition may depend on the properties of the data generation process.
All adaptive variants outperform the SPICE baseline (21.6\%) under a matched update budget, indicating that consensus-gated distillation remains effective even under non-stationary training distributions.
We view adaptive challenger optimization as a promising direction for fully self-contained self-distillation, though further work is needed to develop evaluation protocols and reliability mechanisms suited to non-stationary settings.

\paragraph{Scaling and broader task families}
Scaling to larger models and more diverse document-grounded tasks is a natural next step.
The asymmetric-context setting extends naturally beyond math to any domain where a privileged source document can inform answer generation---including retrieval-augmented generation, tool-use tasks, and agentic settings with privileged environmental state.
However, broader task families will require careful monitoring of document leakage, answer extractability, and consensus calibration.
Understanding how consensus thresholds and rollout budgets should scale with model capability remains an open question.
% ----------------
\section{Conclusion}
We studied self-distillation in a document-grounded question answering setting without verified labels, verifiable rewards, or external graders.
Naive distillation is fragile in self-training because agreement does not imply correctness.
We first use tutor consensus to decide when the model's own supervision is trustworthy; conditioned on those cases, trajectory-level distillation transfers document-grounded reasoning into a document-free student.
Consensus gating alone is sufficient for reliable distillation; adding a sparse correctness reward does not improve student transfer in our setting.
Ablations confirm that the gate is the active ingredient: removing it while keeping all other losses yields performance comparable to unfiltered trajectory distillation.
Empirically, \methodname{} improves held-out student accuracy from 46.0\% to 62.0\%, substantially outperforming answer-only fine-tuning and outcome-based reinforcement learning.

\section*{Acknowledgments}
We thank Avi Schwarzschild for invaluable feedback, discussions, 
and brainstorming throughout this project.

This work was supported by DARPA TIAMAT, the NSF TRAILS Institute (2229885), Coefficient Giving, and Longview Philanthropy.

% \section*{Impact Statement}

% This paper presents work whose primary goal is to advance the field of machine learning by developing more sample-efficient and stable training methods. Our method studies self-improvement in settings without labeled data or external graders, using self-generated signals to support learning.

% As with many advances in machine learning, techniques that reduce reliance on labeled data could lower the cost of training models and broaden access to machine learning technologies. 
% At the same time, such methods could be misused if applied without appropriate safeguards in high-stakes or safety-critical settings. 
% We do not introduce new deployment mechanisms, autonomous decision-making systems, or applications targeting sensitive domains, and our experiments are limited to controlled benchmark settings.

% We believe the ethical considerations associated with this work are consistent with those commonly encountered when advancing general-purpose learning algorithms, and we do not foresee specific negative societal impacts beyond those already well understood in the broader machine learning literature.

\bibliography{refs}
\bibliographystyle{unsrtnat}

\newpage
\appendix
\setcounter{table}{5}
\onecolumn
% \section{You \emph{can} have an appendix here.}

% The $\mathtt{\backslash onecolumn}$ command above can be kept in place if you
% prefer a one-column appendix, or can be removed if you prefer a two-column
% appendix.  Apart from this possible change, the style (font size, spacing,
% margins, page numbering, etc.) should be kept the same as the main body.
% %%%%%%%%%%%%%%%%%%%%%%%%%%%%%%%%%%%%%%%%%%%%%%%%%%%%%%%%%%%%%%%%%%%%%%%%%%%%%%%
% %%%%%%%%%%%%%%%%%%%%%%%%%%%%%%%%%%%%%%%%%%%%%%%%%%%%%%%%%%%%%%%%%%%%%%%%%%%%%%%

\section{Appendix}\label{sec:appendix}

\subsection{Prompts}\label{sec:prompts}

\subsubsection{Student Prompt}

\begin{figure}[ht]
\centering
\begin{minipage}{0.95\linewidth}
\ttfamily\small
Question:\\
\{question\}

\vspace{0.5em}
Solve this step by step.\\
Show your work, then put your FINAL answer in \textbackslash boxed\{\} at the very end.\\
Just answer directly.

\vspace{0.5em}
Solution:
\end{minipage}
\caption{Student prompt used during training and evaluation. The student does not have access to the document.}
\end{figure}

\subsubsection{Tutor Prompt}

\begin{figure}[ht]
\centering
\begin{minipage}{0.95\linewidth}
\ttfamily\small
Document:\\
\{document\}

\vspace{0.5em}
Question:\\
\{question\}

\vspace{0.5em}
Solve this step by step.\\
Show your work, then put your FINAL answer in \textbackslash boxed\{\} at the very end.\\
Do NOT mention the document, passage, or text. Just answer directly.

\vspace{0.5em}
Solution:
\end{minipage}
\caption{Tutor prompt used during training. The tutor has access to the document but is explicitly instructed not to mention it.}
\end{figure}

\subsection{Auxiliary Loss Terms}\label{sec:auxiliary_losses}

Section~\ref{sec:objectives} formally introduces the primary loss terms used in the optimization process.  
In addition to the two distillation loss terms, we experimented with two other terms that can optionally be applied to the total loss.

\paragraph{Consensus-correctness reward}
When enabled, we include a REINFORCE-style objective using a binary pseudo-correctness label $r_{i,j} \in \{0,1\}$ derived from tutor consensus, where $r_{i,j} = 1$ if rollout $j$'s extracted answer matches the consensus for question $i$.
Applied on tutor rollouts under document context:
\begin{equation}
\begin{aligned}
\mathcal{L}_{\text{cons}}(\theta)
&= -\frac{1}{\sum_{i,j} g_i\,L^{(T)}_{i,j}}
\sum_{i,j} g_i\,r_{i,j}
\sum_{t=1}^{L^{(T)}_{i,j}} \tilde{\ell}^{(T)}_{i,j,t}(\theta),\\
\qquad \tilde{\ell}^{(T)}_{i,j,t}(\theta) &\coloneqq \log \pi_\theta\!\left(y^{(T)}_{i,j,t}\,\mid\, y^{(T)}_{i,j,<t},d_i,q_i\right).
\end{aligned}
\end{equation}

\paragraph{KL regularization}
We optionally regularize the student toward a frozen reference policy $\pi_{\text{ref}}$:
\begin{equation}
\begin{aligned}
\mathcal{L}_{\text{KL}}(\theta)
&= \beta\,\frac{1}{\sum_{i,j} g_i\,L^{(S)}_{i,j}}
\sum_{i,j} g_i
\sum_{t=1}^{L^{(S)}_{i,j}} D_{i,j,t}(\theta),\\
\qquad D_{i,j,t}(\theta) &\coloneqq \mathrm{KL}\bigl(\pi_\theta(\cdot\mid y^{(S)}_{i,j,<t},q_i)\,\|\,\pi_{\text{ref}}(\cdot\mid y^{(S)}_{i,j,<t},q_i)\bigr),
\end{aligned}
\end{equation}
where $\beta$ is a tunable coefficient.

\paragraph{Total objective}
The overall training objective is:

\begin{equation}
\mathcal{L}(\theta)
= \lambda_{\text{off}}\,\mathcal{L}_{\text{off}} + \lambda_{\text{on}}\,\mathcal{L}_{\text{on}}
+ \lambda_{\text{cons}}\,\mathcal{L}_{\text{cons}} + \lambda_{\text{KL}}\,\mathcal{L}_{\text{KL}}
\end{equation}

where $\mathcal{L}_{\text{cons}}$ is disabled by default.
The coefficients $\lambda_{\cdot}$ trade off dense trajectory-level learning ($\mathcal{L}_{\text{off}}$, $\mathcal{L}_{\text{on}}$) against the optional sparse correctness reward ($\mathcal{L}_{\text{cons}}$) and stabilization ($\mathcal{L}_{\text{KL}}$), while all terms remain gated by tutor consensus.
We find empirically that the consensus-correctness reward does not improve student transfer in our setting and can reduce student accuracy; consensus-gated trajectory distillation alone provides sufficient supervision (\S\ref{sec:ablations}).

\subsection{Hyperparameters}\label{sec:hyperparameters}

% \paragraph{Training hyperparameters.}
\begin{table}[h]
\centering
\small
\begin{tabular}{ll}
\toprule
\textbf{Hyperparameter} & \textbf{Value} \\
\midrule
Batch size ($n$ questions/step) & 32 \\
Tutor rollouts per question ($k$) & 8 \\
Student rollouts per question ($k$) & 8 \\
Training temperature (tutor/student) & 0.5 \\
Training top-$p$ (tutor/student) & 1.0 \\
Training top-$k$ (tutor/student) & $-1$ \\
Max completion tokens (train rollouts) & 512 \\
Consensus gate (tutor agreement) & $\geq 4/8$ \\
Consensus criterion & Extracted final-answer equivalence \\
Consensus action & Skip question (zero loss) if below threshold \\
Advantage clip ($a$) & $\pm 5$ \\
Off-policy distillation weight ($\lambda_{\text{off}}$) & 1.0 \\
On-policy distillation weight ($\lambda_{\text{on}}$) & 0.1 \\
Consensus reward weight ($\lambda_{\text{cons}}$) & 0 or tuned per ablation \\
KL-to-reference weight ($\lambda_{\text{KL}}$) & .02 \\
Optimizer & AdamW \\
Learning rate & $1 \times 10^{-6}$ \\
Weight decay & 0.01 \\
Gradient clipping & 1.0 \\
\bottomrule
\end{tabular}
\caption{Training hyperparameters used unless otherwise specified.}
\end{table}

% \paragraph{Evaluation hyperparameters.}
\begin{table}[h]
\centering
\small
\begin{tabular}{ll}
\toprule
\textbf{Hyperparameter} & \textbf{Value} \\
\midrule
Evaluation temperature (benchmarks, greedy) & 0 \\
Evaluation temperature (benchmarks, maj@8) & 0.6 \\
Evaluation samples (maj@8 / pass@8) & 8 \\
Eval max completion tokens & 1024 \\
Correctness model for in-domain eval & Qwen2.5-32B-Instruct (consensus) \\
Correctness rollouts (in-domain eval) & 8 \\
Correctness temperature (in-domain eval) & 0.3 \\
Correctness min agreement (in-domain eval) & $\geq 5/8$ \\
Answer extraction & Final boxed answer \\
Equivalence checking & \texttt{math-verify}~\citep{math_verify} \\
\bottomrule
\end{tabular}
\caption{Evaluation hyperparameters used in this submission.}
\end{table}

\subsection{Training Guardrails}\label{sec:guardrails_appendix}
We found the following implementation guardrails critical for stable self-distillation under asymmetric context.

\paragraph{Tokenization and loss accounting}
All losses (distillation, oracle/correctness terms, and any KL regularization) are computed on \emph{completion tokens only}, ensuring that supervision is applied only to model-generated content.
Prompt tokens (document, question, and the \texttt{Solution:} prefix) are masked out of the loss.
We additionally enforce an exact boundary condition: prompts must end with exactly \texttt{Solution:} (no trailing whitespace), so the completion region is unambiguous.

\paragraph{Prompt isomorphism between tutor and student}
Tutor and student prompts share the same structure and delimiter, differing only by the presence of the document for the tutor.
This avoids distribution shift in how the model is trained to begin its completion, isolating the effect of privileged context from prompt-format artifacts.

\paragraph{Formatting, parsing, and rollout validity}
We require each rollout to contain a parsable final answer (the last \verb|\boxed{...}|).
Rollouts missing a boxed answer, suffering truncation before the boxed answer, or failing extraction are treated as \emph{invalid} and receive zero weight in the corresponding loss.

\paragraph{Document leakage prevention}
Because the student will not have document access at test time, we use a two-stage document-mention filter: we drop questions that explicitly reference the document (e.g., "document", "passage", "text"), and we exclude tutor trajectories that mention the document from distillation targets.
This prevents the student from learning document-dependent templates that would be invalid at test time.

\paragraph{Loss interaction rules}
We avoid contradictory gradients by applying each loss only where its supervision signal is meaningful:
(i) off-policy distillation uses only eligible tutor rollouts;
(ii) on-policy distillation uses only valid student rollouts;
(iii) optional oracle/correctness terms may treat malformed outputs as negative examples.

\paragraph{Evaluation hygiene}
We keep evaluation deterministic by using fixed eval IDs, fixed prompts, and a fixed grading pipeline (answer extraction + equivalence checking).
We separately track validity rates (the fraction of rollouts producing a boxed answer) in addition to accuracy, as malformed outputs are a common failure mode in self-distillation.

\subsection{Additional Evaluation Results}\label{app:additional_eval} 
We report supplementary evaluation results under greedy decoding and coverage (pass@8) strategies across all four document-free math benchmarks. 
These complement the maj@8 results in Figure~\ref{fig:main_results}(a).

\begin{figure}[h!]
\centering
\includegraphics[width=\textwidth]{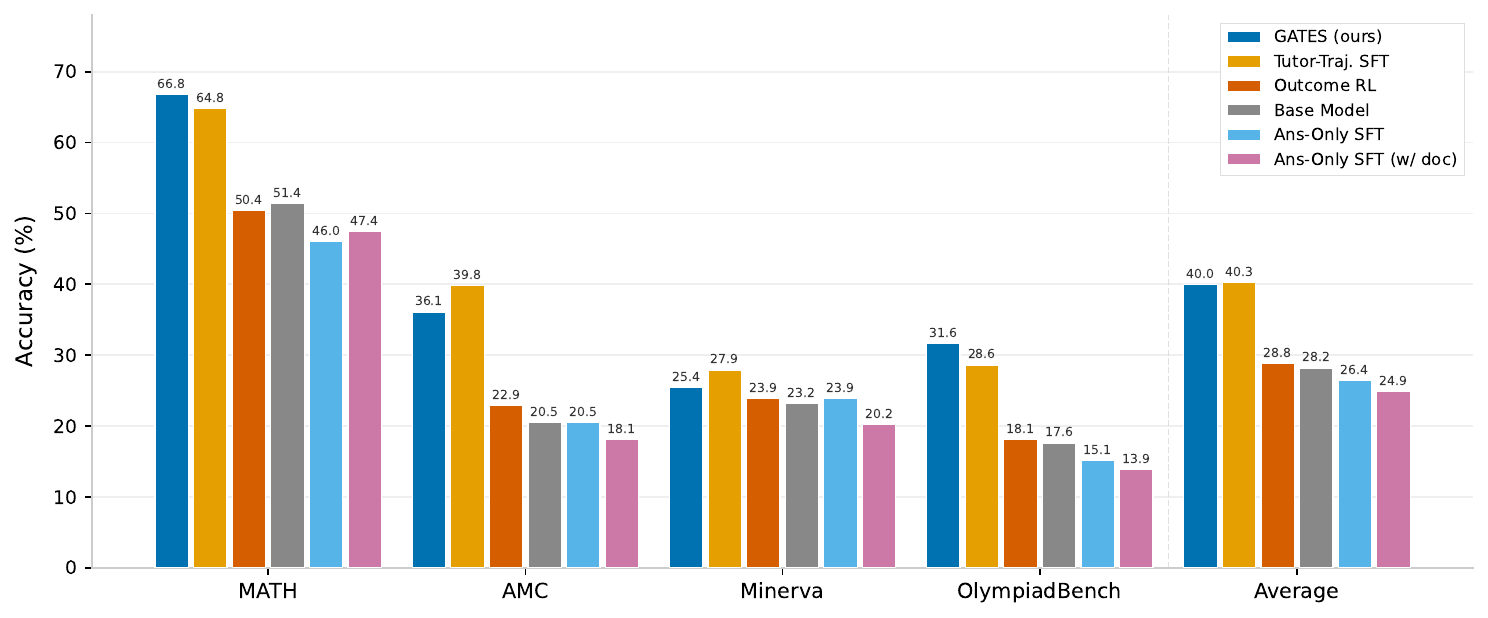}
\caption{Accuracy (\%) under greedy decoding on document-free math benchmarks. 
\methodname{} and Tutor-Trajectory SFT achieve comparable greedy performance, but \methodname{} leads by 3.1 percentage points under maj@8 decoding (Figure~\ref{fig:main_results}a).}
\label{fig:greedy}
\end{figure}

\paragraph{Greedy decoding}
Figure~\ref{fig:greedy} reports accuracy under greedy decoding. 
Under this single-sample metric, \methodname{} and Tutor-Trajectory SFT achieve near-identical average accuracy (40.0\% vs.\ 40.3\%), with Tutor-Trajectory SFT slightly ahead on AMC and Minerva.
The gap between these methods is substantially larger under maj@8 decoding (Figure~\ref{fig:main_results}a), suggesting that consensus gating improves the consistency of correct answers across samples rather than peak single-sample performance.

\begin{figure}[h!]
\centering
\includegraphics[width=\textwidth]{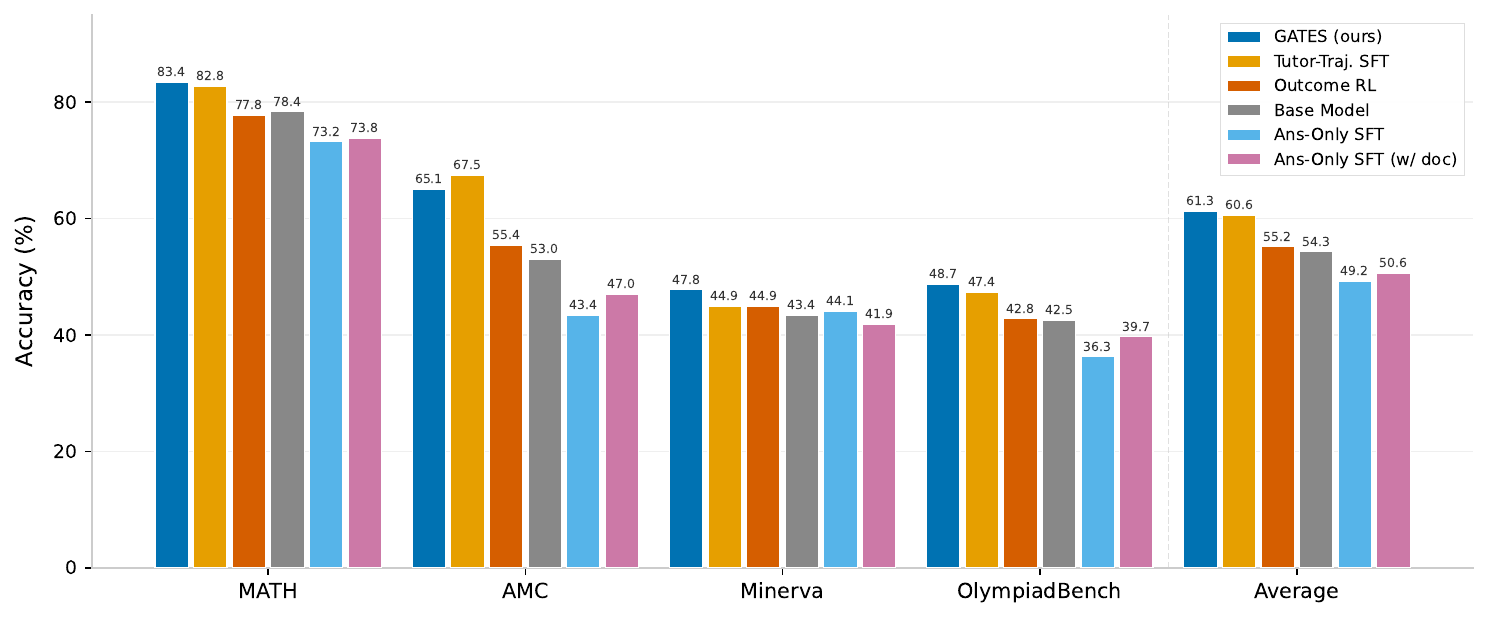}
\caption{Accuracy (\%) under coverage (pass@8) on document-free 
math benchmarks. 
Gaps between methods are smaller than under greedy or maj@8 decoding, consistent with pass@8 measuring an upper bound on model capability.}
\label{fig:pass8}
\end{figure}

\paragraph{Coverage (pass@8)}
Figure~\ref{fig:pass8} reports pass@8, which measures whether at least one of the 8 sampled completions is correct. 
Gaps between methods narrow under this metric, as expected---pass@8 reflects the model's capability ceiling rather than its typical-case accuracy. 
\methodname{} still leads with an average of 61.3\%, compared to 60.6\% for Tutor-Trajectory SFT, 55.2\% for Outcome~RL, and 54.3\% for the base model. 
The smaller margin here indicates that much of the improvement from self-distillation comes from making existing capabilities more reliably accessible, rather than introducing entirely new problem-solving abilities.

\subsection{Tutor Accuracy and the Tutor--Student Gap}\label{sec:student_tutor_gap}

\begin{figure}[h]
\centering
\includegraphics[width=0.85\textwidth]{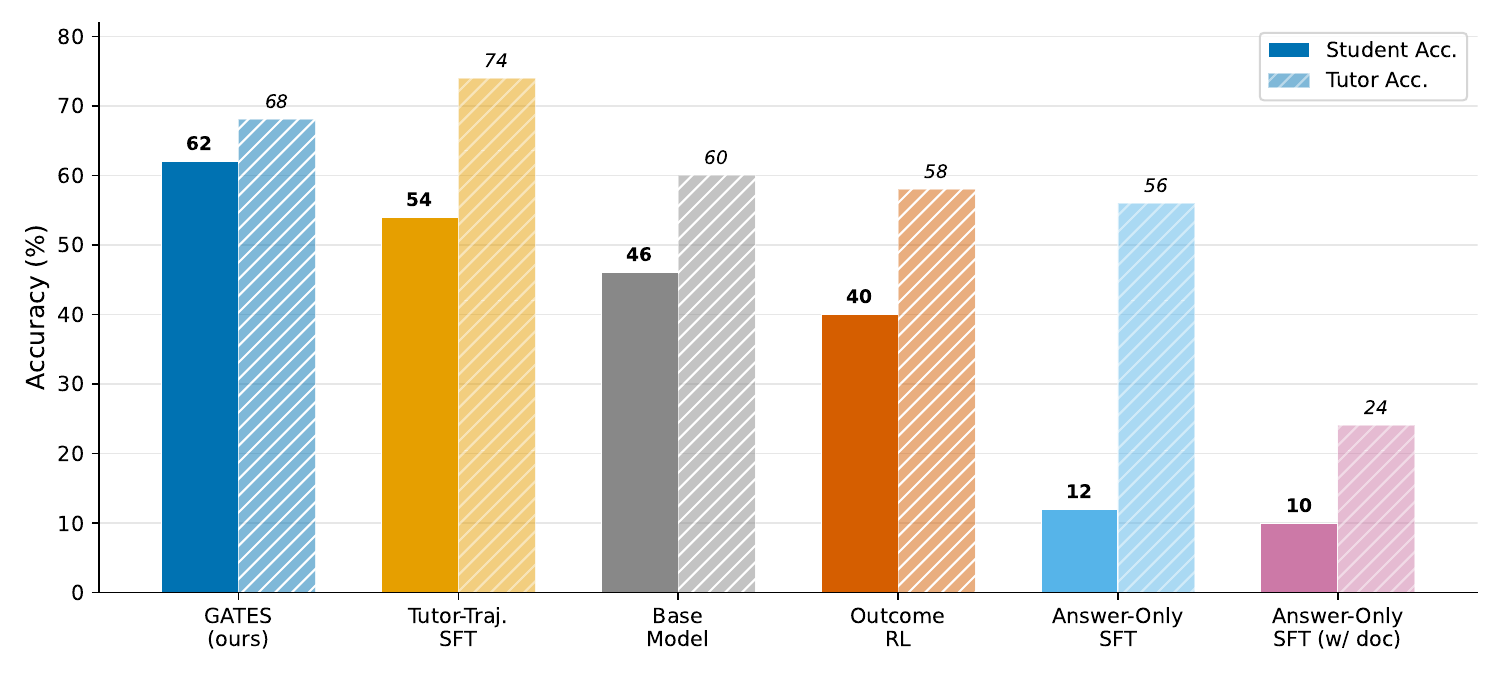}
\caption{Student accuracy (solid) and tutor accuracy (hatched) on the held-out asymmetric evaluation (50 questions, greedy decoding).
Tutor-Trajectory SFT achieves the highest tutor accuracy (74\%) but transfers poorly to the student (54\%),
illustrating that training on unfiltered tutor rollouts inflates tutor performance without improving student internalization.
\methodname{} yields the best student accuracy (62\%) with a narrower tutor--student gap, indicating more effective knowledge transfer via consensus gating.}
\label{fig:appendix_indomain_tutor}
\end{figure}

\begin{figure}[!t]
\centering
\includegraphics[width=0.85\textwidth]{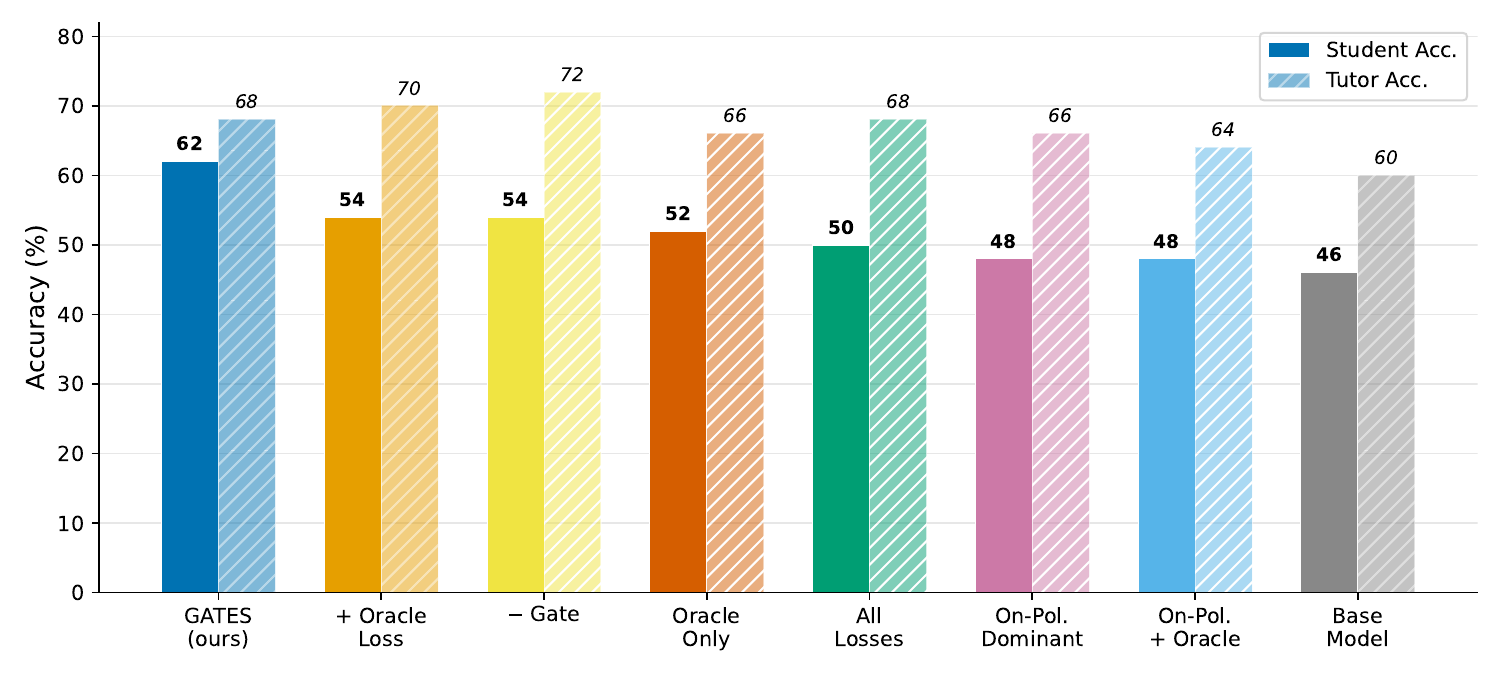}
\caption{Ablation results: student accuracy (solid) and tutor accuracy (hatched) on the held-out asymmetric evaluation (greedy decoding).
Off-policy distillation remains the dominant contributor to student accuracy.
Notably, $-$~Gate achieves the highest tutor accuracy (72\%) but only 54\% student accuracy, mirroring the pattern observed with Tutor-Trajectory SFT and confirming that ungated training inflates tutor performance without improving student transfer.}
\label{fig:appendix_ablation_tutor}
\end{figure}

Figures~\ref{fig:appendix_indomain_tutor} and~\ref{fig:appendix_ablation_tutor} report both student and tutor accuracy on the held-out asymmetric evaluation.
The tutor--student gap reveals how effectively each method transfers document-grounded reasoning to the document-free student.
Tutor-Trajectory SFT achieves the highest tutor accuracy of any baseline (74\%) but only 54\% student accuracy,
producing the widest tutor--student gap in our experiments.
Without consensus gating, the student appears to internalize reasoning patterns that implicitly depend on document access:
the tutor can recover from flawed intermediate steps by re-reading the source, but the student cannot.
Consensus gating filters out these document-dependent trajectories, ensuring the student learns only from reasoning that is self-sufficient.
\methodname{} achieves both the highest student accuracy (62\%) and a narrower tutor--student gap (6 pp vs.\ 20 pp for Tutor-Trajectory SFT),
indicating more effective knowledge transfer.
A similar pattern appears in the ablations:
$-$Gate achieves the highest tutor accuracy of any ablation (72\%) but only 54\% student accuracy,
mirroring Tutor-Trajectory SFT and confirming that ungated training inflates tutor performance without improving student transfer.

\end{document}